\documentclass{article}

\PassOptionsToPackage{numbers}{natbib}

\usepackage[preprint]{nips_2018}

\usepackage[utf8]{inputenc} 
\usepackage[T1]{fontenc}    
\usepackage{hyperref}       
\usepackage{url}            
\usepackage{booktabs}       
\usepackage{amsfonts}       
\usepackage{nicefrac}       
\usepackage{microtype}      
\usepackage{amsmath}
\usepackage{graphicx}
\usepackage{multirow}
\usepackage{multicol}
\usepackage{tabu}
\usepackage[]{algorithm2e}

\def\eg{\emph{e.g. }}
\def\ie{\emph{i.e. }}

\title{How is Contrast Encoded in Deep Neural Networks?}

\author{
  Arash Akbarinia \quad $\&$ \quad
  Karl R. Gegenfurtner \\
  Abteilung Allgemeine Psychologie \\
  Justus-Liebig-Universität \\
  Gießen, Germany \\
  \texttt{$\lbrace$Arash.Akbarinia, Karl.R.Gegenfurtner$\rbrace$@psychol.uni-giessen.de} \\
}

\begin{document}

\maketitle

\begin{abstract}

Contrast is a crucial factor in visual information processing. It is desired for a visual system -- irrespective of being biological or artificial -- to ``perceive'' the world robustly under large potential changes in illumination. In this work, we studied the responses of deep neural networks (DNN) to identical images at different levels of contrast. We analysed the activation of kernels in the convolutional layers of eight prominent networks with distinct architectures (\eg \textit{VGG} and \textit{Inception}). The results of our experiments indicate that those networks with a higher tolerance to alteration of contrast have more than one convolutional layer prior to the first max-pooling operator. It appears that the last convolutional layer before the first max-pooling acts as a mitigator of contrast variation in input images. In our investigation, interestingly, we observed many similarities between the mechanisms of these DNNs and biological visual systems. These comparisons allow us to understand more profoundly the underlying mechanisms of a visual system that is grounded on the basis of ``data-analysis''.

\end{abstract}

\section{Introduction}

The human visual system (HVS) has evolved to be more sensitive to contrast than absolute luminance. Thanks to this intelligent natural selection, we perceive the world steadily despite the huge changes in illumination that we experience throughout the day or across different locations. A similar feature is also vital for machine vision to perform successfully in real-world scenarios, \eg for an autonomous car that is driving in a motorway under diverse lighting conditions.

A recent article \cite{geirhos2017comparing} investigated the impact of image quality on the performance of human and machine vision. A psychophysical experiment was conducted in which the contrast of stimuli was gradually reduced. The results of this study demonstrate that the classification accuracy of \textit{VGG-16} \cite{simonyan2014very} remains on a par with human subjects. A merit that \textit{GoogLeNet} \cite{szegedy2015going} and \textit{AlexNet} \cite{krizhevsky2012imagenet} fall short to meet as their performance significantly deteriorates when contrast of the input image is decreased. This raises an interesting research question regarding the mechanisms involved in \textit{VGG-16} to accomplish this desirable \textit{contrast invariant} behaviour that is absent in the other two networks. Logically, there must be certain operations in place to retain variations of contrast in the input image in order to prevent its propagation to the output of the network.

It is well established, according to numerous physiological studies (e.g. \cite{carandini1997linearity, heeger1992normalization}), that the spike activity of receptive fields (RF) in cells of the visual cortex changes considerably according to the contrast of the presented stimuli. This is also reported (e.g. \cite{shushruth2009comparison,angelucci2013beyond}) in typical centre-surround mechanisms of RFs that have been proposed to play an important role in making a visual system independent of illuminant changes (\ie lightness constancy) \cite{mante2005independence}. While it remains to be seen how exactly this is ``implemented'' in our neuronal system, a naïve equivalent of this for DNNs would be, for example, to change the parameters of a kernel depending on the contrast of neighbouring pixels. A similar approach has been demonstrated to be effective for non-learning algorithms in various computer vision tasks (\eg \cite{arash2017ijcv,akbarinia2017pami}). However, this certainly is not true for \textit{VGG-16} or any other DNNs. Once the network has learnt its parameters, they remain fixed in the test phase. This reinstates the original research question we formulated: \textit{how is contrast invariance achieved in DNNs?}

\begin{figure}[ht]
\centering
\includegraphics[width=\textwidth]{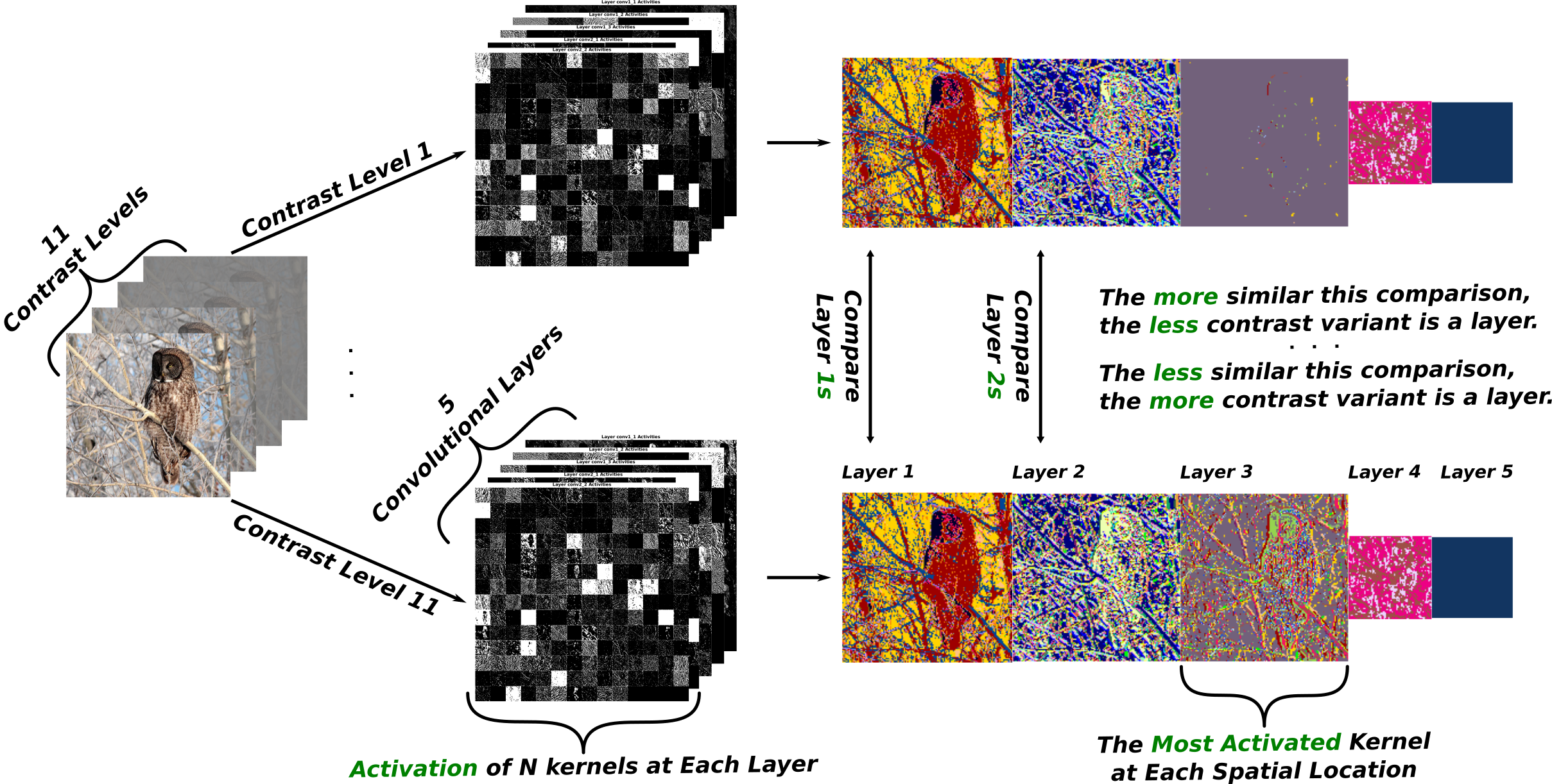}
\caption{Flowchart of the conducted experiment.}
\label{fig:flow}
\end{figure}

In order to answer this question we analysed activation of kernels in convolutional layers of eight prominent networks. Figure \ref{fig:flow} illustrates the flowchart of our experiment. We input each DNN with eleven contrast levels of the same image and measured the activation of all kernels in the first five convolutional layers. For each layer at different spatial locations we retrieved the most activated kernel. Then we computed the proportion of those kernels that remained identical for a given layer across different levels of contrast. This proportion is an indication of whether the behaviour of a layer varies according to the contrast of the stimuli (\ie input image). We compared the corresponding proportions in each layer to examine whether there are layers that account for the contrast invariant behaviour of a DNN. It is worth mentioning that similar procedures are practised commonly to decipher mechanisms of these artificial networks with respect to certain visual features (\eg \cite{kubilius2016deep}).

We conducted an experiment over the entire validation set of \textit{ImageNet} \cite{krizhevsky2012imagenet}. In our analysis a systematic pattern emerged for networks with more than one convolutional layer before the first max-pooling: \textit{the last convolutional layer before the first max-pooling showed more sensitivity to contrast by a large margin}. These networks (\eg \textit{VGG} and \textit{Inception}) also contained their level of accuracy more robustly across different levels of contrast. Others with a max-pooling operator right after the first convolutional layer (\eg \textit{GoogLeNet} and \textit{AlexNet}) did not exhibit this pattern, while performing significantly worse in low contrast images.
%
%
%
%

Given the facts that: (i) DNN is a feedforward model and (ii) the last convolutional layer before the first max-pooling is highly variant to contrast of the input image, we can deduce that as a consequence the entire network achieves contrast invariance. In other words one of the lowest layers in these networks mitigates the variation of contrast, therefore preventing its propagation to higher layers. Conceptually this is in line with architecture of both biological \cite{hubel1962receptive} and artificial networks \cite{lecun2015deep}: lower areas being responsive to basic features (\eg contrast and orientation) while higher areas encode more abstract notions (\eg shapes and objects).

In our investigation we observed a number of curious similarities between the biological and artificial vision. For instance, the invariance to contrast only emerges for networks that have multiple convolutional layers before the first max-pooling, loosely resembling the concept of a cortical area, \eg the primary visual cortex (V1), with its constituting multiple layers, \ie six. Interestingly, in case of biological organisations, it has been hypothesised that contrast is a fundamental independent variable encoded by the early visual system \cite{mante2005independence}. It is all not that surprising to encounter similarities between DNNs and the network inside our brains. Many fundamental hierarchical correspondences between the two have been reported previously \cite{cichy2016comparison}. It seems that both systems act in accordance with the empirical theory of vision \cite{purves2011we}: \textit{inference through successful behaviour}. Both systems have learnt a set of visual features over a period of time (being the evolution or the training phase). Contrast certainly is among the most important low-level features for any visual system. Therefore, we believe our results can shed light on the mechanisms involved around this visual feature: \textit{contrast}.
%
%
%
%
%
%

\section{Experiment}
\label{sec:met}

\subsection{Networks}

We included eight prominent networks of different architectures in our study: \textit{Inception-V3} \cite{szegedy2016rethinking}, \textit{VGG-19} \cite{simonyan2014very}, \textit{VGG-16} \cite{simonyan2014very}, \textit{VGG-16-3C} \cite{he2017channel}, \textit{ResNet-50} \cite{he2016deep}, \textit{ResNet-101} \cite{he2016deep}, \textit{GoogLeNet} \cite{szegedy2015going}, and \textit{AlexNet} \cite{krizhevsky2012imagenet}. For all these DNNs, except \textit{VGG-16-3C}, we used the pretrained networks of the \textit{MathWorks Neural Network Toolbox}. For \textit{VGG-16-3C} we downloaded its \textit{Caffe} version from their \textit{GitHub} repository and imported it to \textit{Matlab} using \textit{importCaffeNetwork} function.

In this article, we restricted the conducted experiments and corresponding analysis to the first five convolutional layers of each network because the shallowest architecture of all (\ie \textit{AlexNet}) only consists of five convolutional layers. In this manner, the comparison among networks would be more consistent. We only studied convolutional layers since they are the pillars of a DNN and carry most of the operations and parameters.

\subsection{Dataset}

We conducted our experiments on a large visual database (\ie \textit{ImageNet} \cite{krizhevsky2012imagenet}) that contains one thousand different categories of objects (\eg plants, animals, cars, bicycles, \emph{etc.}, to name a few). Specifically, we used the validation set of \textit{ImageNet Large-Scale Visual Recognition Challenge} (\textit{ILSVRC 2012}) \cite{russakovsky2015imagenet} with a total number of 50,000 test images (\ie fifty images per object category). It is worth mentioning that all the networks studied in this work have been trained on a subset of \textit{ImageNet} database.

\subsection{Generating images of different contrasts}

Given an input image $I$, we generated eleven images at different levels of contrast, $I_c$, as follows:
\begin{align}
I_c(x,y) = \frac{c}{100} \times I(x,y) + \frac{1 - \frac{c}{100}}{2},
\label{eq:contrast}
\end{align}
where $\left\{x,y\right\}$ are pixel coordinates and $c$ is the contrast level that was set in our experiments to: $c \in \left\{ 1, 3, 5, 7, 10, 13, 15, 30, 50, 75, 100 \right\} \%$. The same procedure has been conducted in the psychophysical experiment of \cite{geirhos2017comparing}, therefore, making our findings comparable.
 
\subsection{Experiment procedure}

We have illustrated the pseudocode of our experiment in Algorithm \ref{alg:over}.
Given a network and a test image, we followed these series of steps:
\begin{enumerate}
	\item Generate eleven versions of the same image at different levels of contrast.
    \begin{enumerate}
    	\item For each contrast image, compute activation of all the kernels in the first five convolutional layers (using \textit{activations} function of \textit{Matlab}).
    	\item At every spatial location keep the most activated kernel.
    \end{enumerate}
    \item In a pair-wise comparison across all levels of contrast, compute the percentage of the most activated kernels in a layer that remain identical at every spatial location.
\end{enumerate}

In this manuscript, we have focused our principal analysis on the most activated kernels for those levels of contrast that a network predicts consistently the same object for a given input image (\eg in both 50\% and 100\% levels of contrast a network predicts an ``owl''). The reason for this choice is that when the prediction of a network is inconsistent under two different levels of contrast, the comparison of the most activated kernels is not very meaningful (\eg in 50\% level of contrast a network predicts a ``cup'' and in 100\% level of contrast an ``owl''). Naturally, as the output of the network belongs to two different classes of object, the most activated kernels would be different as well. (Refer to supplementary materials for a comprehensive analysis of all cases, including those in which the network output is dissimilar in different levels of contrast.)

\begin{algorithm}[ht]
 \SetKwInOut{Input}{inputs}
 \Input{DNN; Image}
 \KwData{$Contrast Levels \gets \left\{ 1, 3, 5, 7, 10, 13, 15, 30, 50, 75, 100 \right\} \%$}
 \ForEach{$c \in Contrast Levels$}{
   $I_c \gets AdjustContrast(Image, c);$ \quad \quad // Following Eq. \ref{eq:contrast}\\
   $Prediction \gets classify(DNN, I_c);$ \quad \hspace{1pt} // Examining whether the network predicts correctly \\
   $Convolutional Layers \gets \left\{ 1, 2, 3, 4, 5 \right\}$ \\
   \ForEach{$l \in Convolutional Layers$}{
     $Features \gets activations(DNN, I_c, l);$ \hspace{1pt} // Computing the activation of all kernels \\
     $Winners \gets argxmax(Features);$ \quad \hspace{5pt} // Retain the most activated kernel at each position \\
   }
 }
 \ForEach{$\left\{c_1,c_2\right\} \in Contrast Levels$}{
   \If{$Prediction_1$  \textbf{equals to} $Prediction_2$}{
     // Compute the percentage of identical kernels in $Winners_1$ and $Winners_2$ \;
   }
 }
 \caption{The detailed procedure of conducted experiment.}
 \label{alg:over}
\end{algorithm}

\section{Results}
\label{sec:res}

\begin{figure}[ht]
\centering
\begin{tabu}{lcc}
\multirow{2}{*}{\includegraphics[width=0.52\textwidth]{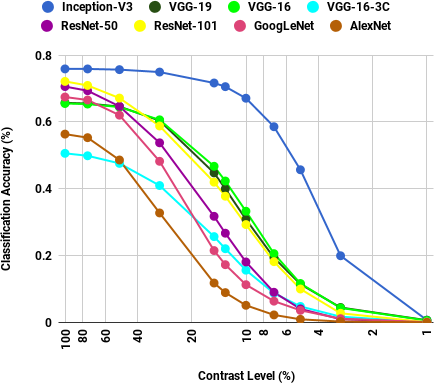}} & \includegraphics[width=0.20\textwidth]{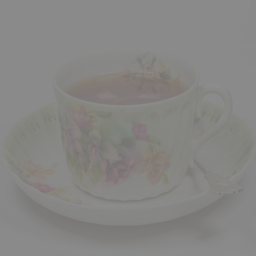} & \includegraphics[width=0.20\textwidth]{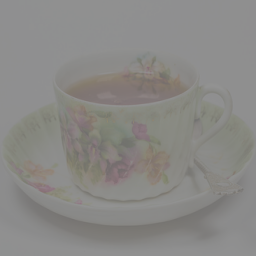} \\
& 10\% Contrast & 15\% Contrast \\
& \includegraphics[width=0.20\textwidth]{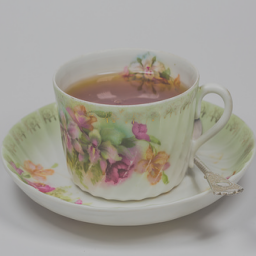} &  \includegraphics[width=0.20\textwidth]{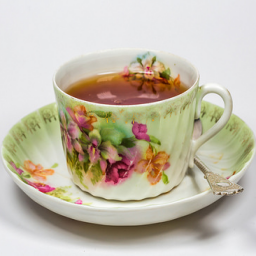} \\
& 50\% Contrast & 100\% Contrast \\
\end{tabu}
\caption{On the left: classification accuracy of networks as a function of image contrast (averaged over 50,000 images of the \textit{ILSVRC 2012} validation set). On the right: an exemplary input image.}
\label{fig:classacc}
\end{figure}

On the right panel of Figure \ref{fig:classacc} we have depicted an exemplary image from the validation set of \textit{ILSVRC 2012} under four different levels of contrast. Naturally, as the contrast of an image is reduced, the classification task becomes increasingly more challenging. Results of a recently reported psychophysical experiment suggest that the performance of humans for the very same task remains almost intact until about 40\% level of contrast \cite{geirhos2017comparing}.

On the left panel of the Figure \ref{fig:classacc} we have reported the classification accuracy of eight networks under eleven levels of contrast. Each point is the average accuracy over 50,000 images. It is worth emphasising that in this study we are not investigating which network obtains the best performance in comparison to the others, whereas we are interested in understanding the behaviour of a network as the contrast level is reduced (an intra-comparison instead of inter-comparison).

If the classification accuracy of a network drops sharply as the contrast level is reduced, this implies the performance of that network is \textit{contrast variant}, an undesirable behaviour. Contrary to that, if a network retains its own peak accuracy in low contrast inputs, this network posses this desired behaviour of \textit{contrast invariance}. According to this criteria, we can observe that \textit{Inception-V3} exhibits the greatest invariance to contrast of the input images, while \textit{AlexNet} performing the worst.

\begin{figure}[ht]
\centering
\begin{tabu}{cc}
\includegraphics[width=0.48\textwidth]{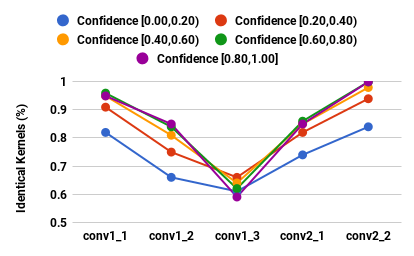} & \includegraphics[width=0.48\textwidth]{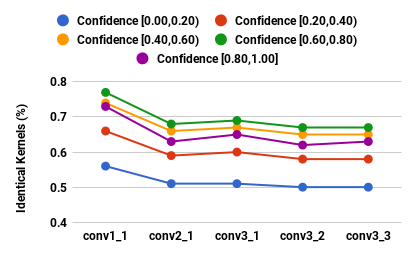} \\
\textit{Inception-V3} & \textit{AlexNet} \\
\includegraphics[width=0.48\textwidth]{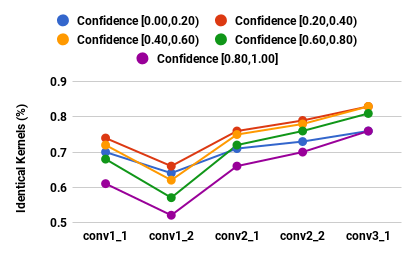} & \includegraphics[width=0.48\textwidth]{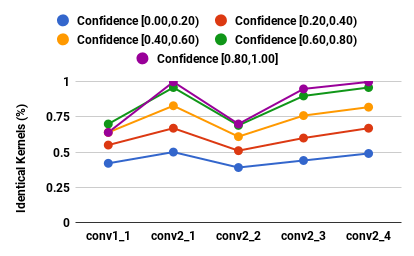} \\
\textit{VGG-19} & \textit{ResNet-101} \\
\end{tabu}
\caption{Portion of the most activated kernels for the first five convolutional layers that remained identical across all levels of contrast. Each series represents different confidence levels of a network.}
\label{fig:bestwo}
\end{figure}

In Figure \ref{fig:bestwo}, for four networks, we have reported the portion of the most activated kernels that remain identical across different levels of contrast (refer to the supplementary materials for the same analysis of all other networks). Each point represents variation in the output of a layer throughout all contrast trials. If this figure is 1 for a layer, this means at each spatial location always the same kernel remains the most activated one, irrespective of contrast of the input image. Contrary to that, if this figure is 0 for a layer, it means that the most activated kernel always changes according to contrast of the input image.

The results of \textit{Inception-V3} is reported on the top left panel of Figure \ref{fig:bestwo}. Let us remind ourselves that the architecture of this network consists of three convolutional layers prior to the first max-pooling. We can observe that \textit{conv1\_3} has the lowest percentage of identical kernels by a large margin. The results of \textit{VGG-19} is reported on the bottom left panel of Figure \ref{fig:bestwo}. The smallest portion of identical kernels belongs to \textit{conv1\_2} (the last convolutional layer prior to the first max-pooling).

On the top right panel of Figure \ref{fig:bestwo} we have reported the results of \textit{AlexNet}. Let us remind ourselves that in this architecture there is a max-pooling after the first convolutional layer. We can observe that the curvatures are almost flat. On the bottom right panel of Figure \ref{fig:bestwo} we have reported the results of \textit{ResNet-101}. We cannot detect any clear patterns between these two networks that perform a max-pooling operation after the first convolutional layer.

In Table \ref{tab:all} we have reported the percentage of identical kernels for all studied networks (similar to Figure \ref{fig:bestwo} however irrespective of the confidence of the network). Double vertical lines in this Table represents a max-pooling layer. All networks with more than one convolutional layer prior the first max-pooling (the first four rows) share two identical patterns:
\begin{enumerate}
	\item The lowest percentage of the identical kernels always belongs to the convolutional layer right before the first max-pooling (bold figures).
    \item The highest percentage of the identical kernels always belongs to the fifth convolutional layer (the last column of the Table).
\end{enumerate}

Contrary to that, we cannot observe any clear pattern for all those networks that perform a max-pooling operation right after the first convolutional layer (the last four rows).

\begin{table}[ht]
\centering
\caption{Portion of the most activated kernels for the first five convolutional layers that remained identical across all contrast levels. Double vertical lines indicate a max-pooling operator.}
\setlength{\tabcolsep}{16pt}
\renewcommand{\arraystretch}{1.2}
\begin{tabular}{|l|ccccc|}
\cline{2-6}
\multicolumn{1}{l|}{} & \multicolumn{5}{c|}{\textbf{Convolutional layers ordered from lower to higher}} \\ \hline
\textbf{\textit{Inception-V3}} & 0.94 & 0.82 & \multicolumn{1}{c||}{\textbf{0.61}} & 0.84 & \textit{0.98} \\ \hline
\textbf{\textit{VGG-19}} & 0.67 & \multicolumn{1}{c||}{\textbf{0.58}} & 0.70 & \multicolumn{1}{c||}{0.74} & \textit{0.79} \\ \hline
\textbf{\textit{VGG-16}} & 0.69 & \multicolumn{1}{c||}{\textbf{0.59}} & 0.68 & \multicolumn{1}{c||}{0.75} & \textit{0.78} \\ \hline
\textbf{\textit{VGG-16-3C}} & 0.71 & 0.68 & 0.71 & \multicolumn{1}{c||}{\textbf{0.65}} & \textit{0.74} \\ \hline \hline
\textbf{\textit{ResNet-50}} & \multicolumn{1}{c||}{0.70} & 0.91 & 0.63 & 0.55 & 0.91 \\ \hline
\textbf{\textit{ResNet-101}} & \multicolumn{1}{c||}{0.63} & 0.92 & 0.65 & 0.87 & 0.92 \\ \hline
\textbf{\textit{GoogLeNet}} & \multicolumn{1}{c||}{0.77} & 0.60 & \multicolumn{1}{c||}{0.68} & 0.62 & 0.62 \\ \hline
\textbf{\textit{AlexNet}} & \multicolumn{1}{c||}{0.68} & \multicolumn{1}{c||}{0.60} & \multicolumn{1}{c||}{0.61} & \multicolumn{1}{c||}{0.59} & 0.59 \\ 
\hline
\end{tabular}
\label{tab:all}
\end{table}

\section{Discussion}
\label{sec:dis}

\subsection{Performance of networks}

The results of our experiment across different levels of contrast (Figure \ref{fig:classacc}) demonstrate that four networks (\ie \textit{Inception-V3}, \textit{VGG-19}, \textit{VGG-16}, and \textit{VGG-16-3C}) largely retain their classification accuracy down to 40\% level of contrast, which was reported to be also the case for human subjects \cite{geirhos2017comparing}. This implies that these networks are \textit{invariant to contrast} of the input images, similar to our very own visual system. This desired feature is more pronounced in case of \textit{Inception-V3} that maintains its performance as far as 15\% level of contrast.

Contrary to this, the performance of the other four networks (\ie \textit{ResNet-50}, \textit{ResNet-101}, \textit{GoogLeNet}, and \textit{AlexNet}) deteriorates rapidly as contrast of the input images is reduced. For instance, although at 100\% level of contrast the classification accuracy of \textit{GoogLeNet} is superior to \textit{VGG-16},  at 40\% contrast \textit{VGG-16} is significantly better than \textit{GoogLeNet}. This becomes even more noticeable at lower levels of contrast (compare the pink and green curves in Figure \ref{fig:classacc}). A similar trend occurs if we compare \textit{AlexNet} to \textit{VGG-16-3C} (the brown and cyan curves, respectively).

The distinction between the two regimes of networks cannot be traced back to their learning procedures, as all have been trained with a subset of the \textit{ImageNet} dataset with no particular data augmentation regarding the contrast of input images. The difference cannot arise from their corresponding depths either. \textit{GoogLeNet} is substantially deeper than the \textit{VGG} networks. The topology of the networks can neither explain this distinction. The architecture of \textit{Inception-V3} is of a directed acyclic graph (DAG) while the \textit{VGG} family is serial. The number of operations executed neither appears to be decisive: \textit{Inception-V3} performs much less operations in comparison to \textit{ResNet-101}.

The only evident pattern is that the architecture of all \textit{contrast invariant} networks consists of more than one convolutional layer prior to the first max-pooling operation: \textit{Inception-V3} has three convolutional layers before the first max-pooling, \textit{VGG-19} and \textit{VGG-16} two, and \textit{VGG-16-3C} four. Contrary to this, all other four networks perform a max-pooling operation after their fist convolutional layer. This loosely, yet interestingly, resembles cortical areas of biological visual systems that are composed of multiple layers. For instance, the primary visual cortex (V1) of humans is divided into six layers \cite{douglas1998neocortex}.
%
%

\subsection{Analysis of layers}

A clear pattern emerges for the \textit{contrast invariant} networks in our analysis of the consistency of the most activated kernels across different levels of contrast (Table \ref{tab:all}): the lowest portion of identical kernels always belongs to the convolutional layer before the first max-pooling, with a large margin. For instance, within \textit{Inception-V3} the difference between \textit{conv1\_3} to all other layers is more than 20\%. Similarly, this figure is about 10\% lower for \textit{conv1\_2} of \textit{VGG-19} and \textit{VGG-16} with respect to the other layers of those networks. This margin is less pronounced among the convolutional layers of \textit{VGG-16-3C} perhaps because \textit{conv1\_4} of this network considers a minimum spatial neighbourhood.

This observed pattern implies a potential hypothesis regarding the underlying strategy of these networks: \textit{in order to achieve contrast invariance, the lower convolutional layers (before the first max-pooling) mitigate the variation of contrast in the input image}. Given the fact that the architecture of DNNs is feedforward, a consequence of this hypothesis would be that higher layers become less sensitive to contrast. We exactly observe this consequence in Table \ref{tab:all}: the highest convolutional layer we studied (\ie the fifth) always obtains the largest value of identical kernels within each of these four networks. This is a coherent strategy: contrast is a low-level feature and its variation should not be propagated to higher layers of hierarchy as their purpose is detecting more generic features. 
%
%
%
%

\begin{figure}[ht]
\centering
\begin{tabu}{cc}
\includegraphics[width=0.48\textwidth]{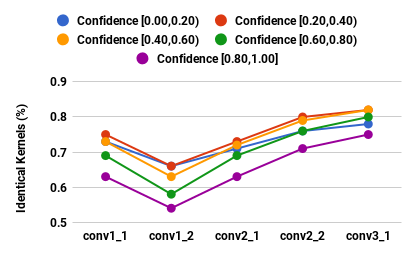} & \includegraphics[width=0.48\textwidth]{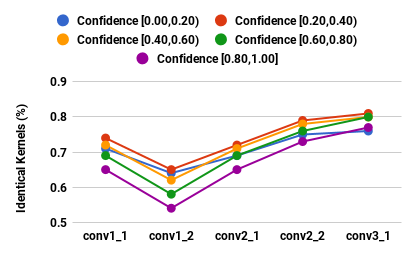} \\
\textit{VGG-16} Correct Classification & \textit{VGG-16} Incorrect Classification \\
\end{tabu}
\caption{Portion of the most activated kernels for the first five convolutional layers that remained identical across all levels of contrast. Each series represents different confidence levels of a network.}
\label{fig:corvsincor}
\end{figure}

This pattern is not determined by whether a network predicts an object correctly according to the ground-truth (compare the left and right panels of Figure \ref{fig:corvsincor} for \textit{VGG-16} and refer to supplementary materials for all other networks). At the same time, although, this pattern is not conditioned to the confidence of a network, it is more noticeable at higher levels of confidence. For instance, we can observe that the margin between \textit{conv1\_2} to all other layers of \textit{VGG-16} is substantially larger when the confidence of the network is between 80-100\% in comparison to 0-20\% (compare the purple and blue curves in both panels of Figure \ref{fig:corvsincor}). This is also the case for \textit{conv1\_3} of \textit{Inception-V3} and \textit{conv1\_2} of \textit{VGG-19} (compare the purple and blue curves in their corresponding panels of Figure \ref{fig:bestwo}).

The reason that this pattern is influenced by the confidence of the network and not by its correctness is rather logical. As far as a network is concerned, when it has a high confidence it thinks that its prediction is correct, regardless of what the ground-truth states. The fact that this pattern becomes more pronounced in higher levels of confidence suggests that this strategy (\ie mitigating the variation of contrast before the first max-pooling) is truly aiding the network to reach a more ``correct'' decision.

\subsection{Contrast comparison}

\begin{figure}[ht]
\centering
\begin{tabu}{cc}
\includegraphics[width=0.48\textwidth]{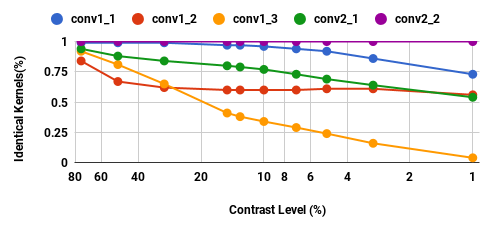} & \includegraphics[width=0.48\textwidth]{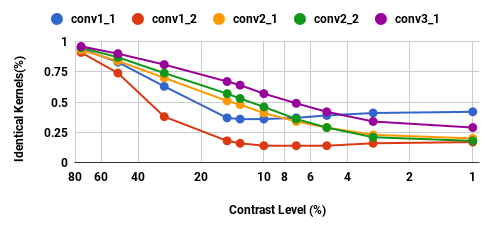} \\
\textit{Inception-V3} & \textit{VGG-19} \\
\end{tabu}
\caption{A pair-wise comparison between 100\% level of contrast to all other contrast levels.}
\label{fig:pw}
\end{figure}

In order to scrutinise more thoroughly our inferences, we analysed the portion of identical kernels in a pair-wise comparison between 100\% level of contrast to all other levels of contrast. We have reported this analysis for two networks in Figure ~\ref{fig:pw} (refer to supplementary materials for the other networks). A value of 1 for the percentage of identical kernels implies that given the input image in one level of contrast, the most activated kernels at every spatial locations remain identical to those at 100\% level of contrast (\ie the activity of a layer is not affected by contrast of the input image). Contrary to this, as this figure becomes smaller it means that the behaviour of a layer is influenced more by contrast of the input image. We should expect to observe a sharper decline in percentage of identical kernels for the layer we are proposing that mitigates the impact of contrast (\ie the last convolutional layer before the first max-pooling) and a slower decline for other layers.

Indeed this pattern emerges in Figure ~\ref{fig:pw}. For instance, we can observe that in case of \textit{Inception-V3} at 75\% level of contrast, all five convolutional layers obtain a high percentage of identical kernels. This is natural since the difference between an image at 80\% and 100\% levels of contrast is minimal, therefore the activity of kernels remains similar. As level of contrast is reduced, this figure drops more sharply for \textit{conv1\_3} (the yellow curve) in comparison to all the others. This corroborates our proposal that this layer acts as a mitigator of contrast in input images. For example, the last convolutional layer we studied, \textit{conv2\_2}, has completely a flat line (the purple curve). Interestingly, this invariation is also true for \textit{conv1\_2} (the red curve) that at the start has a lower percentage of identical kernels in comparison to \textit{conv1\_3}. Let us remind ourselves that the classification accuracy of \textit{Inception-V3} does not decrease substantially down until 10\% level of contrast (Figure \ref{fig:classacc}). At this level of contrast more than two thirds of the winner kernels undergo a change in layer \textit{conv1\_3}. This suggests a possible explanation for why the network still can classify objects correctly in low contrast images.

We can observe a similar set of patterns for all the other \textit{contrast invariant} networks. For instance, in case of \textit{VGG-19}, the percentage of the most activated kernels drops more sharply also for the last convolutional layer before the first max-pooling, \textit{conv1\_2} (refer to the red curve on the right panel of Figure \ref{fig:pw}). This is evident to a greater extent down until 30\% level of contrast in which the classification accuracy of the network remains intact (Figure \ref{fig:classacc}); and it is less pronounced at very low levels of contrast (\ie less than 5\%). However, since the classification accuracy of \textit{VGG-19} is very low at those levels of contrast, therefore comparison across layers is not really informative.

\section{Conclusion}

In this article we studied the behaviour of eight prominent DNNs under the variation of contrast in input images. We conducted our investigation on a benchmark image dataset of object recognition (\textit{ImageNet}). We computed the classification accuracy of each network under eleven levels of contrast. The results of our experiments demonstrate that half of the studied networks (\eg \textit{Inception}) exhibit the desired \textit{contrast invariant} feature, similar to what it has been reported previously for humans \cite{geirhos2017comparing}. The performance of other networks (\eg \textit{GoogLeNet}) drops sharply as contrast of the input image is reduced. The only evident pattern to distinguish the two regimes is whether their architecture consists of more than one convolutional layer prior to the first max-pooling.

We further investigated the underlying mechanisms of this feature by analysing the most activated kernels of the first five convolutional layers. We measured the portion of the winner kernels that change as contrast of the input stimuli varies. The results of our analysis reveal that in case of the \textit{contrast invariant} networks the last convolutional layer before the first max-pooling always shows the greatest alteration to contrast of the input image, while the smallest changes belong to the fifth convolutional layer (the highest layer studied in our experiment). This suggests that one layer at lower hierarchy absorbs the variation of contrast in the input images, therefore preventing its propagation to the output of the network.

One major purpose of contrast gain control mechanisms in the primary visual cortex (V1) is to construct a normalised response independent of its contrast \cite{wilson2014configural}. The exact mechanisms and operations involved to achieve this feature remains to be discovered. However, in general we know that the visual response properties of cortical neurons arise to a great extent as a consequence of the organisation and function of their connections with other neurons \cite{callaway2004cell}. Therefore, it can be hypothesised that neurons in higher cortical areas -- responsible of detecting objects -- are largely invariant to contrast as a results of pooling from a different set of winner neurons. In other words, this feature is not grounded on an intrinsic operation of theirs, but rather owing to extrinsic stream of data from lower cortical areas.
%
%

Whether this is true for biological neural networks ought to be examined, however, we observed this strategy in artificial DNNs. A sequence of convolutional layers increases the spatial region over which visual information is integrated, while pooling reduces the dimensionality of its representation. Contrast is standard deviation of intensity relative to the mean of a region. Accordingly, those networks with more than one convolutional layer before the first max-pooling ground their contrast representation on a richer set of details. In other words, this strategy allows those networks to encode contrast of the training images (environment) more precisely. Interestingly, V1 contrast adaptation mechanisms have been reported to match the statistics of the environment (natural images) \cite{mante2005independence}.
%
%

\subsubsection*{Acknowledgements}

This project was funded by the Deutsche Forschungsgemeinschaft SFB/TRR 135. A preliminary version of our experiments has been presented in abstract form in European Conference on Visual Perception (ECVP) 2018 \cite{arashkarlecvp2018}.

\small

\bibliographystyle{plain}
\bibliography{nips18}

\end{document}